\pdfoutput=1
\documentclass[11pt]{article}

\usepackage{authblk}
\usepackage{mtsummit25}

\usepackage{pdflscape} 
\usepackage{rotating}
\usepackage{adjustbox}

\usepackage{times}
\usepackage{latexsym}

\usepackage{multicol}
\usepackage{subcaption}

\usepackage[T1]{fontenc}
\usepackage[utf8]{inputenc}

\usepackage{microtype}

\usepackage{inconsolata}

\usepackage{graphicx}

\usepackage{float}
\title{MultiSlav: Using Cross-Lingual Knowledge Transfer to Combat the Curse of Multilinguality}

\author[1]{Artur Kot}
\author[1]{Mikołaj Koszowski}
\author[1]{Wojciech Chojnowski}
\author[1]{Mieszko Rutkowski}
\author[2]{\authorcr Artur Nowakowski}
\author[2]{Kamil Guttmann}
\author[2]{Mikołaj Pokrywka}

\affil[1]{Machine Learning Research Allegro.com, \{name\}.\{surname\}@allegro.com}
\affil[2]{Laniqo.com, \{name\}.\{surname\}@laniqo.com}

\begin{document}
\maketitle
\begin{abstract}
Does multilingual Neural Machine Translation (NMT) lead to \emph{The Curse of the Multlinguality} or provides \emph{the Cross-lingual Knowledge Transfer} within a language family? In this study, we explore multiple approaches for extending the available data-regime in NMT and we prove cross-lingual benefits even in \textbf{0-shot} translation regime for low-resource languages. With this paper, we provide state-of-the-art open-source NMT models for translating between selected Slavic languages. We released our models on the HuggingFace Hub\footnote{Link: \url{https://hf.co/collections/allegro/multislav-6793d6b6419e5963e759a683}} under the \emph{CC BY 4.0} license. 
Slavic language family comprises morphologically rich Central and Eastern European languages. Although counting hundreds of millions of native speakers, Slavic Neural Machine Translation is under-studied in our opinion. Recently, most NMT research focuses either on: high-resource languages like English, Spanish, and German - in WMT23 General Translation Task \cite{kocmi-etal-2023-findings} 7 out of 8 task directions are from or to English; massively multilingual models covering multiple language groups; or evaluation techniques. 
\end{abstract}

\section{Introduction} 

In the literature, we can find 2 seemingly contradictory observations about multilingual models: (1) adding more languages to NLP models will lead to \emph{Cross-lingual Knowledge Transfer} increasing the quality of the model, notably for low-resource languages and primarily for related or geographically co-located languages \cite{koloski-etal-2022-thin, adelani-etal-2022-masakhaner}; (2) adding more languages to the model may lead to \emph{The Curse of the Multilinguality}, reducing the quality of the model, especially for high-resource languages \cite{conneau-etal-2020-unsupervised}. The rule of thumb is: that only languages from the same language group (and written in the same script) should increase the quality of the model. However, we are not aware of any study validating or disproving this claim for Slavic languages.

In this paper, we provide a study of the application of a Multilingual NMT approach to the group of low- and mid-resource (defined in \autoref{sect:data}) languages represented by selected Latin-script Slavic languages: Czech, Polish, Slovak, and Slovene.

\begin{figure*}
\subfloat[\label{fig1:a}][Bi-Directional Data (baseline 63M examples)]
{\includegraphics[width=.45\linewidth]{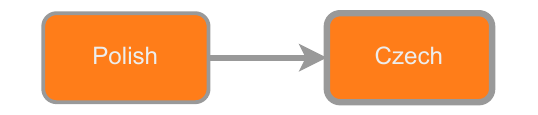}}\hfill
\subfloat[\label{fig1:b}][Pivot Slavic Data (+55M examples)]
{\includegraphics[width=.45\linewidth]{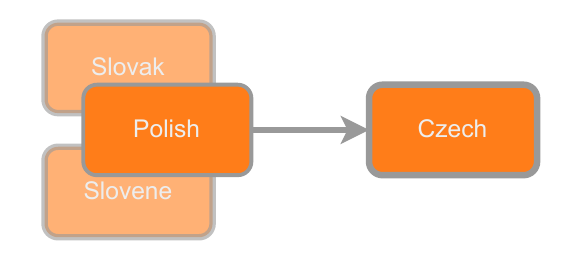}}\par
\subfloat[\label{fig1:c}][Multilingual Slavic Data (+122M)]
{\includegraphics[width=.45\linewidth]{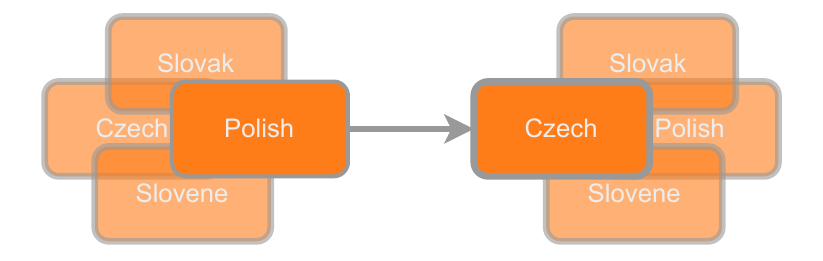}}\hfill
\subfloat[\label{fig1:d}][Multilingual Slavic + English Data (+515M)]
{\includegraphics[width=.45\linewidth]{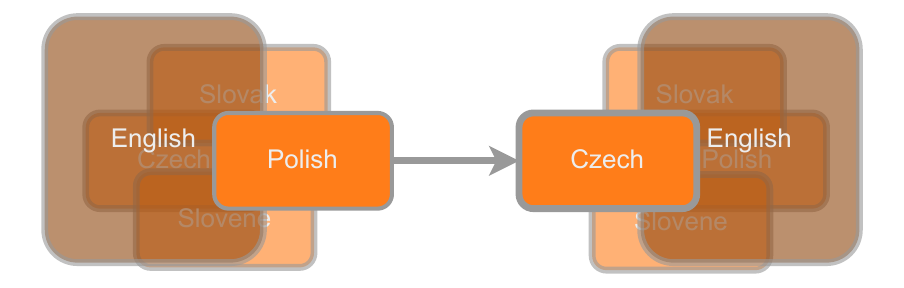}}
 \caption{Strategies for increasing the data-regime without decreasing the quality of the model illustrated by example of translating from Polish to Czech language. In parenthesis we show how many data points were added compared to baseline.}
\label{fig1:approaches}
\end{figure*}

We explore the extension of this group by adding the high-resource language - English. The English language is culturally influential in the modern era whilst also providing access to a large number of parallel examples (bitext) for selected languages, increasing the open-source pool by a factor of {\small{$3$}}. Explored strategies are presented in \autoref{fig1:approaches}.

\begin{table}
\centering
\begin{tabular}{rlr}
\hline
\textbf{ } & \textbf{Language Pairs} & \textbf{Data size}\\
\hline
\verb|1| & $Czech \leftrightarrow Polish$ & $63M$ \\
\verb|2| & $Czech \leftrightarrow Slovak$ & $30M$ \\
\verb|3| & $Czech \leftrightarrow Slovene$ & $25M$ \\
\verb|4| & $Polish \leftrightarrow Slovak$ & $26M$ \\
\verb|5| & $Polish \leftrightarrow Slovene$ & $23M$ \\
\verb|6| & $Slovak \leftrightarrow Slovene$ & $18M$ \\\hline
\verb|7| & $English \leftrightarrow Czech$ & $151M$ \\
\verb|8| & $English \leftrightarrow Polish$ & $150M$ \\
\verb|9| & $English \leftrightarrow Slovak$ & $52M$ \\
\verb|10| & $English \leftrightarrow Slovene$ & $40M$ \\
\hline
 & \emph{$Slavic$ $directions$ } & $185M$ \\
 & \emph{$All$ $directions$} & $578M$ \\\hline
\end{tabular}
\caption{Size of open-source training bitext for each pair of languages in Millions of parallel sentences. Size is counted after filtering and deduplication.}
\label{tab:datasets-sizes}
\end{table}

In our study: (1) we evaluate several multilingual translation scenarios: Bi-Directional models, Multi-way Multilingual models ($Many2Many$) \cite{firat-etal-2016-multi} and Pivot models \cite{kim-etal-2019-pivot, utiyama-isahara-2007-comparison} and evaluate the quality of the methods for each of the selected Slavic languages; (2) we evaluate how adding English language to the selected group impacts performance of the models; (3) we investigate the impact of \emph{the Cross-lingual Knowledge Transfer} in a narrow language group.

The results of this study confirm \emph{the Cross-lingual Knowledge Transfer} hypothesis for the translation between Slavic languages. Indeed, multilingual training increases the quality of the lower data-regime direction (e.g. Slovak to Slovene), even in the \emph{directional zero-shot} \cite{johnson-etal-2017-googles} regime and on the language pair not present in the training data.

\section{Related work}
\label{sect:related_work}

The underlying system architecture of most of the commercial MT vendors is closed and unknown but released information suggests heavy use of transformers architecture \cite{NIPS2017_3f5ee243} and $Many2Many$ translation models \cite{johnson-etal-2017-googles}. Meta released multiple multilingual MT models, starting from mBART \cite{liu-etal-2020-multilingual-denoising}, M2M \cite{10.5555/3546258.3546365}, NLLB \cite{nllb2022} and most recently SeamlessM4T \cite{seamlessm4t2023}.

There are also fully open-source initiatives led by the University of Helsinki centered around OPUS corpora collection \cite{tiedemann-2012-parallel} with multiple releases of MT models \cite{tiedemann-thottingal-2020-opus, tiedemann-2020-tatoeba}.

Prior research on the \emph{cross-lingual knowledge transefer} in NMT was conducted for Indic \cite{10.1145/3587932}, and Turkic \cite{mirzakhalov-etal-2021-evaluating} language families.

Moreover, Large Language Models (LLMs) have recently entered the scene of Machine Translation. Proprietary LLMs often outperform custom translation models on the high-resource languages \cite{kocmi-etal-2023-findings}, but still lag behind classical solutions in the mid-, and low-resource regime \cite{hendy2023good, zhu2023multilingual}. For extensive reviews on the multilingual machine translation methodologies see \citet{kocmi-2021} and \citet{10.1145/3406095}.

\section{Data}
\label{sect:data}

\subsection{Data Sources}

Training datasets were downloaded via MTData library \cite{gowda-etal-2021-many}, see details in \autoref{app:detailed-data-sources} and in \autoref{tab:data-details-app}. The aggregated sources, languages supported, and size of each corpus can be found in \autoref{tab:datasets-sizes}. For evaluation, we use the parallel dataset from Flores 101 - dev \cite{goyal-etal-2022-flores}, which contains 997 sentences translated into multiple languages including all 5 languages in our scope. 

\subsection{Data Filtering}
\label{subsect:filtering}

Firstly, we normalize text by removing special characters, unifying quotations and whitespaces, and applying the Unicode NFKC (Normalization Form Compatibility Composition) normalization.

Then we filter out potentially misaligned sentences using the following text-based features:
(1) Levenshtein distance between source and target sentences; (2) sentence length in characters; (3) sentence length in amount of tokenized words; (4) FastText language detection \cite{joulin2016fasttext}\footnote{FastText model version: LID.176}; (5) Poisson-based log-probability for sentence length ratios \cite{koszowski-etal-2021-allegro}; (6) mismatched numbers; (7) ratio of digits to other characters; (8) average length of tokenized words; (9) maximum length of the longest word in the sentence; (10) alphabet-based non-whitelist character ratio to the rest of characters. 
Duplicate training pairs are removed if either side, source sentence, or target sentence, is already present in the dataset.

\section{Experimental Setup}
\label{sect:experiments}

In this section, we describe tokenizer training, special tokens for language hinting, and model architecture used for training models.

\begin{table*}
\begin{adjustbox}{max width=0.95\textwidth}
\centering

\begin{tabular}{l l l l}
\hline
Model & Type & Directions supported & Size \\\hline
$baseline$ $(ours)$ & bi-directional &  All 20 directions & 10 models $ 209 M$ each \\
\hline
Google Translate & March 2024  & All 20 directions & N/A \\
$PaLM-2$ & March 2024  & All 20 directions & N/A \\
$GPT-3.5$ & March 2024  & All 20 directions & N/A \\
\hline
$M2M-100$ & Many2Many & All 20 directions & $1.2B$ \\
$NLLB-200$ & Many2Many & All 20 directions & $1.3B$ \\
$OPUS-MT$  $Sla-Sla^*$ & Many2Many & 12 directions, missing $SLK$ pairs & $64M$\\
$OPUS-MT$  $SK-EN^*$ & Bi Directional & 2 directions: $SLK \leftrightarrow ENG$ & \\
\hline
$MultiSlav$ $(ours)$ & Many2Many & 12 directions, missing $ENG$ pairs & $242 M$ \\
$MultiSlav _{+ENG}$ $(ours)$ & Many2Many & All 20 directions & $258 M$ \\
\hline
$P4_{POL}$ $(ours)$ & Pivot via Polish & 12 directions, missing $ENG$ pairs & 2x$242 M$ \\
$P5_{ENG}$ $(ours)$ & Pivot via English & All 20 directions & 2x$ 258 M$ \\
$P5_{CES}$ $(ours)$ & Pivot via Czech & All 20 directions & 2x$258 M$ \\
\hline

\end{tabular}
\end{adjustbox}
\caption{Available solutions, baseline, and proposed methods.* - due to missing $SLK$ the results for OPUS-MT are reported in \autoref{app:detailed-results}}
\label{tab:model-info}
\end{table*}

\subsection{Tokenizer}

We used the SentencePiece unigram model \cite{kudo-richardson-2018-sentencepiece} as a tokenizer. Based on our experiments with tokenizer sizes (see \autoref{app:tokenzier-sizes}), we concluded that $16k$ tokens are a sufficient size to cover each language. Therefore, for bidirectional, four-language, and five-language models, we use $32k$, $64k$, and $80k$ vocabulary sizes, respectively.

The tokenizers are trained on subsets of the entire training corpus. In each case, we sampled around $40M$ sentences total. Duplicate sentences were removed before the sampling. English is the dominant language in the corpora. Among the Slavic corpora - Slovene and Slovak are noticeably smaller than Czech and Polish. To mitigate the potential impact of data imbalance per language, we experimented with sampling an equal amount of data for each language and proportionally sampling the percentage of the training set, across the languages. 

We chose \emph{the equal sampling} for all models. This strategy prevents the over-representation of English sentences.  More details on the tokenizer training experiments can be found in \autoref{app:tokenizer-sampling}. During the tokenizer training, we added special tokens to identify languages of the translation direction, as described in \autoref{sub:lang-tok}.

\subsection{Language tokens}
\label{sub:lang-tok}

To ensure the correct output language, multilingual models require indicating the target language. We achieve this by prepending the source sentences with \emph{special tokens} of the target language. They are constructed as \verb|>>X<<|, where \verb|X| stands for the lowercase ISO-639-3 three-letter language code, as described in \citet{tiedemann-thottingal-2020-opus}. For example, for translating from Polish to Czech, we add \verb|>>ces<<| to the source sentence. We did not observe any performance differences between models using only the target language token, and both source and target language tokens. More details regarding this choice are described in the \autoref{app:lang-tok}.

\subsection{Architecture}

The Encoder-Decoder post-layer normalization transformer \cite{NIPS2017_3f5ee243} is the base architecture for all of the trained models. We use three-way tying of embedding matrices between source, target, and output. This architecture is used for Bi-Directional, Many2Many ($MultiSlav$), and Pivot (both $One2Many$ and $Many2One$) models. 

All of the models trained by us have the same number of non-embedding parameters, but the total number of parameters differs due to the different vocabulary sizes used. 
Models are trained using the \mbox{MarianNMT} library \cite{junczys-dowmunt-etal-2018-marian}. 
To indicate the direction of the translation we use special language-hinting tokens (\emph{lang-tokens}) providing context for the model. In total, we trained 18 models.
We report model sizes in the \autoref{tab:model-info}. See \autoref{app:hyper-params} for the training details and hyper-parameter choice.

\begin{figure}
    \centering
    \includegraphics[width=1.0\linewidth]{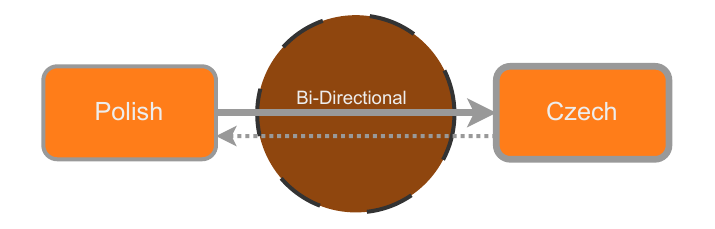}
    \caption{\emph{Bi-directional Model} translates in both directions between 2 languages.}
    \label{fig:bi-da-diagram}
\end{figure}

\subsection{Bi-Directional Models}

As the $baselines$ we trained bi-directional models. 
They are capable of translating in both directions within language pairs, e.g., translating both from Polish to Czech ($POL\rightarrow CES$) and from Czech to Polish ($CES\rightarrow POL$) with one Bi-directional Czech \& Polish model ($CES\leftrightarrow POL$), see \autoref{fig:bi-da-diagram}. 
We trained 10 Bi-Directional models for each combination of the supported languages.

\subsection{Pivot Model}

\begin{figure*}
\centering
    \centering
    \includegraphics[width=.95\linewidth]{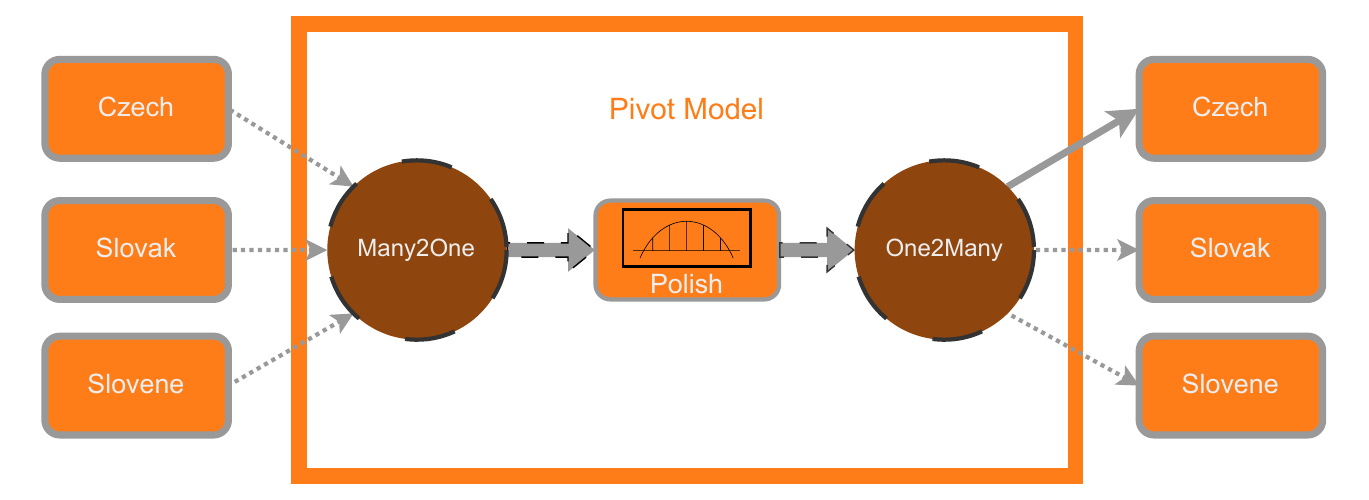}
    \caption{Pivot system uses 2 models: (1) translates from multiple languages to \emph{Bridge Language} and second from \emph{Bridge Language} to multiple languages - effectively translating between all supported languages.}
    \label{fig:p4-diagram}
\end{figure*}

As a \emph{Pivot Model}, we understand the system of 2 NMT models translating: (1) from multiple languages via the \emph{Bridge Language} ($Many2One$) and (2) from one \emph{Bridge Language} to multiple languages ($One2Many$), see \autoref{fig:p4-diagram}. Each of them is trained separately. It allows us to increase the \emph{Bridge Language} data examples in the training set. It may also increase the fluency (correctness of the target language) in the $Many2One$ model and the accuracy (understanding of the source language) in the $One2Many$ model. Using languages from the same language group could potentially utilize \emph{the Cross-lingual Knowledge Transfer}.
The inference in the \emph{Pivot Model} consists of 3 cases: 
 (1) translating TO the \emph{Bridge Language}, the source sentence is translated via the $Many2One$ model;
 (2) translating FROM the \emph{Bridge Language}, the source sentence is translated via the $One2Many$ model;
 (3) otherwise, the source sentence is translated to the \emph{Bridge Language} through the $Many2One$ model into the \emph{Bridge Sentence}, then the \emph{Bridge Sentence} is passed to the $One2Many$ model to translate into the target language (pivot translation through \emph{Bridge Language}).

Firstly, we trained \emph{4 Slavic languages} pivot ($P4$) via Polish ($P4_{POL}$). Then to expand the dataset size, we chose to train the model for \emph{Pivot 5 Languages} ($P5$) with English as the \emph{Bridge Language} ($P5_{ENG}$), which is the high-resource language.
Unlike English, Slavic languages are morphologically rich; this information may be lost while using English as \emph{Bridge Language}. To quantifiably evaluate this risk, we trained a model for $P5$ with pivot through Czech ($P5_{CES}$). 

By using \emph{Pivot Model} for $P5$ we reduce the number of needed models covering all 20 directions from 10 bi-directional $baselines$ to only 2 models ($P5 Many2One$ and $P5 One2Many$). In total, we trained 3 \emph{Pivot Models}: $P4_{POL}$, $P5_{ENG}$ and $P5_{CES}$. Each one is a system of 2 separate models ($Many2One$ and $One2Many$).

\subsection{Multilingual Models}

\begin{figure*}
    \centering
    \includegraphics[width=.75\linewidth]{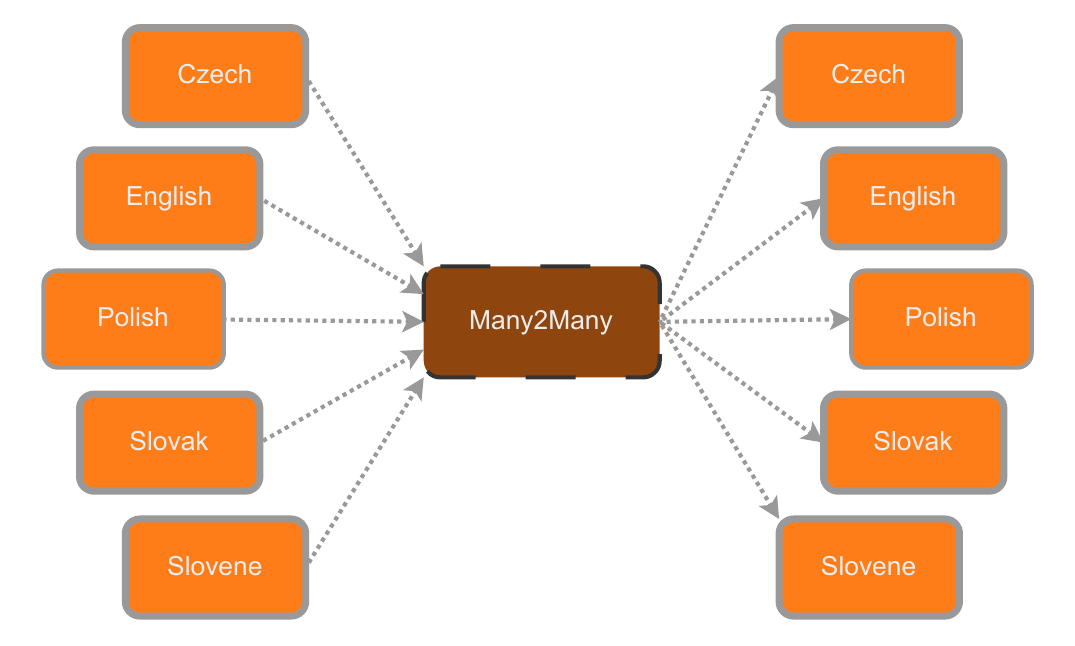}
    \caption{\emph{Multilingual Model} directly translates between all supported languages.}
    \label{fig:mutlislav-diagram}
\end{figure*}

Utilizing \emph{Pivot Models} allowed us to increase the data points for the \emph{Bridge Language}. This potentially improved \emph{Bridge Language} fluency and accuracy. However, \emph{Pivot Models} did not use any available bitext for other translation directions between supported languages. We predict this could lead to a decrease in the translation quality in different directions. Additional problems may arise from the accumulation of errors through multi-step translation.
To reduce those risks and utilize bitext for all directions, we chose to train \emph{Multilingual Models}.
\emph{Multilingual Models} ($Many2Many$) are translating between multiple languages. We trained two such models: Multi-Slavic 4 language model ($MultiSlav$) translating between Czech, Polish, Slovak, and Slovene and Multi-Slavic 5 language model ($MultiSlav _{+ENG}$) translating between Czech, English, Polish, Slovak, and Slovene - see \autoref{fig:mutlislav-diagram}.

\section{Results}
\label{sect:results}

To assess the translation quality, we use lexical metric {\verb|chrF|}\footnote{Exact version and configuration chrF2: nrefs:1 case:mixed eff:yes nc:6 nw:0 space:no version:2.3.1 from SacreBLEU library} \cite{popovic-2015-chrf} and neural metric {\verb|COMET|}\footnote{Used COMET model: Unbabel/wmt22-comet-da} \cite{rei-etal-2022-comet}. \verb|COMET| and \verb|chrF| show better correlation to expert evaluation than historically used \verb|BLEU| \cite{papineni-etal-2002-bleu}.
The primary metric used for the analysis in this section is \verb|COMET|, due to having the highest correlation to human experts out of the above-mentioned metrics \cite{freitag-etal-2022-results}.

\autoref{tab:results-avg} shows the results averaged for all \emph{All 20 directions} and \emph{12 Slavic directions} respectively. 12 directions are defined as the Cartesian product of 4 Slavic languages $\{CES, POL, SLK, SLV\}$ and 20 directions as the Cartesian product of 5 languages: 4 Slavic and English.  Additionally, in \autoref{tab:high-low-resource}, we provide results for: a) the highest-resource Slavic pair ({\small{$CES \leftrightarrow POL$}}); and b) for the lowest-resource Slavic pair ({\small{$SLK \leftrightarrow SLV$}}).
For completeness, we provide detailed results in \autoref{app:detailed-results}, for each direction, and each of metrics: \verb|chrF|, \verb|COMET| and \verb|BLEU|.

Aggregated results show that the closed commercial translation system (represented by Google Translate) and the LLMs in the zero-shot approach (represented by PaLM-2 \cite{anil2023palm} in the \textit{text-bison@002} and ChatGPT-3.5 in the \textit{turbo-0125} versions) achieve high performance on automatic metrics. However, data in \autoref{app:detailed-results} shows that commercial models score higher in some directions (e.g. {\small{$SLV \rightarrow POL$}}, {\small{$+2.7$}} \verb|COMET|) but are comparable to the $baselines$ in other directions (e.g. {\small{$SLK \rightarrow CES$}} {\small{$-0.1$}} \verb|COMET|). Open-source massively multilingual models \verb|M2M-100| and \verb|NLLB-200| provide translations on average worse than baselines ({\small{$-0.7$}} and {\small{$-1.3$}} \verb|COMET| points, respectively), while being 2-5 times larger. 

Models from \verb|OPUS-MT| (variants \verb|Sla-Sla| and \verb|SK-EN|) did not cover all selected directions, due to this fact, we had to exclude them from aggregated results. However, by analyzing specific translation directions, we can see competitive results in some directions. Overall \verb|OPUS-MT|, while being smaller models often showed better results than $baselines$ for English directions (e.g., scoring the highest of all open-source models in {\small{$CES \rightarrow ENG$}}). \verb|OPUS-MT| results for directions between Slavic languages are significantly lower.

\begin{table*}[h]
\centering
\begin{adjustbox}{max width=0.95\textwidth}
\begin{tabular}{l| c c | c c}
  & \multicolumn{2}{c|}{a) $All$ $directions$ Avg(std)} & 
    \multicolumn{2}{c}{b) $Slavic$ $directions$ Avg(std)} \\
 & $chrF$ & $COMET$ & $chrF$ & $COMET$ \\
\hline
$GoogleTranslate$ & $\underline{57.5}(5.9)$ & $90.5(1.4)$ & $\underline{53.6}(2.6)$ & $91.0(0.8)$ \\
$PaLM-2$\textsuperscript{*} & $57.2(5.7)$ & $\underline{90.7}(1.6)$ & $53.5(2.5)$ & $\underline{91.3}(1.1)$ \\
$ChatGPT-3.5$\textsuperscript{*} & $55.1(5.8)$ & $89.8(1.4)$ & $51.6(3.2)$ & $90.4(1.0)$ \\
\hline
$M2M-100$ & $54.1(5.2)$ & $88.7(1.9)$ & $51.3(3.4)$ & $89.9(1.3)$ \\
$NLLB-200$ & $53.5(6.3)$ & $89.0(1.3)$ & $49.4(2.7)$ & $89.4(1.2)$ \\
$Seamless-M4T$ & $51.2(8.0)$ & $84.5(4.9)$ & $45.8(4.5)$ & $82.0(4.7)$ \\
\hline
$baseline$ & $54.8(5.3)$ & $88.6(1.9)$ & $51.8(3.4)$ & $89.8(1.5)$ \\
\hline
$P4_{POL}$ & - & - & $51.0(2.1)$ & $89.4(0.9)$ \\
$P5_{ENG}$ & $55.0(5.6)$ & $88.5(1.2)$ & $51.5(3.0)$ & $89.1(0.9)$ \\
$P5_{CES}$ & $54.5(5.4)$ & $88.7(2.2)$ & $51.7(3.5)$ & $90.0(1.5)$ \\
$MultiSlav$ & - & - & $\textbf{52.2}(3.3)$ & $90.2(1.3)$ \\
$MultiSlav_{+ENG}$ & $\textbf{55.2}(5.2)$ & $\textbf{89.2}(1.9)$ & $\textbf{52.2}(3.3)$ & $\textbf{90.4}(1.2)$ \\
\end{tabular}
\end{adjustbox}
\caption{Results for a) all directions and b) Slavic directions. Standard deviation (std) is calculated between the results of different language pairs. \underline{Underlined} are the best results, \textbf{bolded} are the best open-source results. \textsuperscript{*} - for a couple of examples LLMs "\textit{refused}" to provide translation.}
\label{tab:results-avg}
\end{table*}

\begin{table*}[h]
\centering
\begin{adjustbox}{max width=0.95\textwidth}
\begin{tabular}{l| c c | c c}
  & 
    \multicolumn{2}{c|}{a) {\small{$CES \rightarrow POL$}} / {\small{$POL \rightarrow CES$}}} &  
    \multicolumn{2}{c}{b) {\small{$SLK \rightarrow SLV$}} / {\small{$SLV \rightarrow SLK$}}} \\
 & $chrF$ & $COMET$ & $chrF$ & $COMET$ \\
\hline
$GoogleTranslate$ & $\underline{51.6}/50.1$ & $\underline{91.0}/91.0$& $\underline{56.9}/\underline{55.5}$ & $\underline{90.5}/91.1$ \\
$PaLM-2$\textsuperscript{*} & $51.5/\underline{50.2}$ & $\underline{91.0}/\underline{91.6}$& $54.2/55.1$ & $89.3/\underline{91.6}$ \\
$ChatGPT-3.5$\textsuperscript{*} & $49.2/47.8$ & $89.8/90.6$& $55.1/51.8$ & $90.3/89.6$ \\
\hline
$M2M-100$ & $48.0/47.7$ & $89.0/89.6$& $55.0/52.8$ & $89.6/90.1$ \\
$NLLB-200$ & $47.3/46.7$ & $88.9/89.4$& $52.0/50.4$ & $88.8/89.4$ \\
$Seamless-M4T$ & $43.5/41.2$ & $80.9/79.6$& $48.8/46.0$ & $82.9/81.0$ \\
\hline
$baseline$ & $49.2/48.5$ & $89.4/90.0$& $55.4/52.5$ & $89.4/89.1$ \\
\hline
$P4_{POL}$ & $\textbf{49.5}/48.5$ & $89.6/90.2$& $53.2/51.1$ & $88.4/88.5$ \\
$P5_{ENG}$ & $48.7/48.3$ & $89.0/89.0$& $54.6/52.9$ & $88.5/88.9$ \\
$P5_{CES}$ & $49.0/48.6$ & $89.6/90.3$& $55.3/53.0$ & $89.8/89.8$ \\
$MultiSlav$ & $49.2/48.7$ & $89.7/90.2$& $\textbf{55.7}/\textbf{53.6}$ & $90.1/\textbf{90.2}$ \\
$MultiSlav_{+ENG}$ & $49.3/\textbf{48.9}$ & $\textbf{89.8}/\textbf{90.4}$& $\textbf{55.7}/53.3$ & $\textbf{90.2}/\textbf{90.2}$ \\
\end{tabular}
\end{adjustbox}
\caption{Results for a) highest-resource Slavic pair and b) lowest-resource Slavic pair. \underline{Underlined} are the best results, \textbf{bolded} are the best open-source results. \textsuperscript{*} - for a couple of examples LLMs "\textit{refused}" to provide translation.}
\label{tab:high-low-resource}
\end{table*}

\subsection{Baselines}

$Baselines$ bi-directional models proved to be highly competitive to Massively Multilingual models like \verb|M2M-100| or \verb|NLLB-200|. 

\subsection{Commercial methods}

LLMs showed the best overall quality of translations.
The difference to bi-directional models is $+2.0$ \verb|COMET| for 20 directions, and $+1.5$ \verb|COMET| for 12 Slavic directions. The gap highly depends on source and target language: LLMs score $-0.2$ \verb|COMET| on $CES \leftrightarrow SLK$ in respect to baseline. We see similar discrepancies for all methods. For detailed results refer to \autoref{app:detailed-results}.
However, due to the nature of "ensuring safety" - LLMs "refused" to translate some examples which are related to controversial subjects. For more information on the topic, refer to \autoref{app:llm-nontranslate}. 
\verb|Google Translate| provided very close results to LLMs ($-0.2$ \verb|COMET| 20 directions and $-0.3$ \verb|COMET| in 12 Slavic directions) without skipping controversial examples. 

\subsection{Pivot}

Pivot Models did not show overall significant improvements over the $baseline$, on average best Pivot model scoring $+0.1$ \verb|COMET| on all 20 directions and $+0.2$ \verb|COMET| on 12 Slavic directions, often scoring worse.
Notably, if we split pivot into Many2One and One2Many models, each one of $Many2One P4_{POL}$, $One2Many P4_{POL}$, $Many2One P5_{ENG}$, $One2Many P5_{ENG}$, $Many2One P5_{CES}$, and $One2Many P5_{CES}$ was always better than a $baseline$ in both translating to and from the \emph{Bridge Language}. For example $P5_{CES}$ was better than $baseline$ in translating $Any \leftrightarrow CES$, however worse in translating $POL \leftrightarrow SLV$.
Lower quality may stand from accumulating errors in each pass and lack of direct training bitext in that direction.

\subsection{MultiSlav}

Multilingual Many2Many approach of $MultiSlav$ and $MultiSlav_{+ENG}$ showed the most considerable increase in the automatic metrics over the \verb|baseline|; $MultiSlav_{+ENG}$ scoring $+0.6$ \verb|COMET| in both: All 20 directions and 12 Slavic directions. 
The \verb|COMET| score of $MultiSlav_{+ENG}$ improved in 19 out of 20 directions, $SLK \rightarrow CES$ did not change quality; in terms of \verb|chrF| $MultiSlav_{+ENG}$ improved in 18/20 directions, except for $-0.4$ \verb|chrF| in $ENG \rightarrow SLK$ and did not change for $ENG \rightarrow SLV$.
$MultiSlav$ also improved in \verb|COMET| score in 11/12 direction, $SLK \rightarrow CES$ did not change; in \verb|chrF| $MultiSlav$ improved in 11/12 directions, except for $-0.2$ \verb|chrF| in $CES \rightarrow POL$.
$MultiSlav_{+ENG}$ did not show any significant improvement over $MultiSlav$ in 12 Slavic directions in \verb|COMET|.

\section{Knowledge transfer}

To estimate if \emph{the Cross-lingual Knowledge Transfer} occurs in Slavic languages, we trained $baseline$ models, 3 pivot models, and 2 $Many2Many$ models. All 3 pivot models showed increased quality in terms of \verb|COMET| score of translation to and from \emph{Bridge Language}. Multilingual $MultiSlav$ and $MultiSlav_{+ENG}$ models universally improved over $baselines$. \emph{the Curse of Multilinguality} did not occur, neither in $MultiSlav$ nor after adding English data.

However, to prove \emph{the Cross-lingual Knowledge Transfer} we set up the experiment for \textbf{directional zero-shot}. By directional zero-shot, we understand training the model in multiple directions with missing one (or several) directions and evaluating it on missing directions, e.g. evaluating model on direction {\small{$CES \rightarrow POL$}} after training model on {\small{$[CES \leftrightarrow SLK, POL \leftrightarrow SLK]$}} - if model shows the quality comparable to $baselines$ it must have inferred knowledge of \emph{Czech} language accuracy from {\small{$CES \rightarrow SLK$}} and knowledge of \emph{Polish} language fluency from {\small{$SLK \rightarrow POL$}}.

As missing directions, we chose directions from the lowest resource pair: {\small{$SLK \rightarrow SLV$}} and {\small{$SLV \rightarrow SLK$}}.

\subsection{Zero-shot Slovak $\leftrightarrow$ Slovene}
\label{sec:zero-ablation}

We trained 3 additional $Many2Many$ variants of 4 Slavic $MultiSlav$ model: (1) excluding data for {\small{$SLK \rightarrow SLV$}} direction, (2) excluding data for {\small{$SLV \rightarrow SLK$}} direction, (3) excluding data for both {\small{$SLK \rightarrow SLV$}} and {\small{$SLV \rightarrow SLK$}} directions.

Table~\ref{tab:0-shot-results} presents results of the \emph{directional zero-shot} experiment. Each model improved over the $baselines$. This finding proves that the Multilingual model, trained within a language group may still provide a competitive solution for low-resource directions - even if training data for that specific translation direction is unavailable.

Additionally, we observed that model trained without opposite direction (trained for {\small{$SLK \rightarrow SLV$}} without {\small{$SLV \rightarrow SLK$}} and vice versa) in both cases achieves even better results. Further analysis would fall outside of the scope of this study, however, understanding if this is a common occurrence or not, would be an interesting future research.

\begin{table}[h]
\centering
\begin{adjustbox}{max width=0.48\textwidth}
\begin{tabular}{l| c c}
 % & \multicolumn{2}{c}{\small{$SLK\rightarrow SLV / SLV\rightarrow SLK$}} \\
 &  \multicolumn{2}{c}{{\small{$SLK\rightarrow SLV / SLV\rightarrow SLK$}}} \\
 & $ChrF$ & $COMET$ \\
 \hline
$Baseline$ & $55.4/52.5$ & $89.4/89.1$ \\
 \hline
$MultiSlav$ & $55.7/53.4$ & $90.1/90.0$ \\
- exclude \small{$SLK\rightarrow SLV$} & $55.4/\textbf{53.6}$ & $90.1/\textbf{90.2}$ \\
- exclude \small{$SLV\rightarrow SLK$} & $\textbf{55.9}/52.9$ & $\textbf{90.2}/89.9$ \\
- exclude both  & $55.5/53.1$ & $90.0/89.9$ \\
\end{tabular}
\end{adjustbox}
\caption{0-shot ablations for $MultiSlav$.}
\label{tab:0-shot-results}
\end{table}

\section{Conclusions}
\label{sect:conclusions}

In the case of studied languages, the multilingual approach of extending the data-regime proved to have a positive impact for each tested variation. We did not observe any drawbacks that would be potentially brought by the hypothesis of \emph{"The Curse of Multilinguality"}. Even the extension of data by including English language pairs either helped or did not affect the results on pairs of Slavic languages.

In \autoref{sec:zero-ablation} we proved that our models exhibit knowledge transfer within Slavic language pairs - the quality of the translation of $MultiSlav_{excl: SLK \leftrightarrow SLV}$ exceeded $Baseline_{SLK \leftrightarrow SLV}$ on both directions.

The small amount of closely related languages used in one Multilingual model can be a better quality solution overall, and be a convenient model in terms of technical deployment. We recommend referring to the \autoref{app:detailed-results} before choosing a solution for a specific translation direction - $MultiSlav_{+ENG}$ being the most versatile.

\section{Future Work}

In the future study, we want to continue researching cross-linguality. This study did not take into account counter-examples, i.e., using geographically co-located languages outside of a single-family (e.g., Romanian, Hungarian, and Turkish for MultiSlav) or using different language families to extend $SLK \leftrightarrow SLV$ data regime. Another interesting subject would be extending the list of Slavic languages by Cyrillic script languages (e.g., Belarusian, Russian, Serbian, and Ukrainian) - this could show if knowledge transfer manifests only in same-script scenarios or is inherent to in-language-family scenarios despite different alphabets.

\section*{Limitations}

Training multilingual models "\emph{from scratch}" presented in our study requires computational resources, including GPUs and large amounts of RAM, taking up to a week of 4x NVIDIA A100 GPU time. Depending on the region, this also may lead to large emissions. To mitigate this impact, we released our models under the \emph{CC BY 4.0} license, hopefully reducing the need to pretraining those models from scratch by the community.
Our analysis is focused on a single general-domain validation set - \emph{FLORES-101}, results may vary based on the specific domain. 
$P5_{ENG}$ uses English (morphologically limited language) to translate between morphologically rich languages. Among other limitations, this may lead to the loss of explicit gender information and result in reproducing biases from the training data. 
The datasets used were not filtered based on gender, cultural, or racial bias. Users should take that into account.

\bibliography{anthology,mtsummit25,custom}

\newpage
\appendix

\section{Hyper-parameters and training setup}
\label{app:hyper-params}

We list the hyper-parameter values used for training in \autoref{tab:hyper-params-used}. We did not perform the full hyper-parameter tuning, but we experimented with increasing the number of layers of the encoder and decoder to 10. Larger models did not increase the performance and led to instability in training, therefore we used the default value of 6 layers each. For the training procedure, we used Adam optimizer \cite{kingma2014adam} with parameters $\beta_1=0.9$, $\beta_2=0.98$, $\epsilon=10^{-9}$, $\mathtt{lr}=0.0002$ with the linear learning rate warmup for 8000 steps followed by the inverse square root decay. 

Our models were trained on NVIDIA A100 or V100 cards, depending on their availability in the cloud environment we used. The specific number of tokens varied a bit between batches due to the `mini-batch-fit` algorithm we used to fully utilize the requested amount of vRAM on GPUs. For all the trainings, we used the same effective workspace of 128GB using equivalent combinations: the number of GPUs, \texttt{workspace}, and \texttt{optimizer-delay} parameters. For baselines, we utilized full-precision training (float32) for the rest of the models we used mixed-precision, which doubled our effective batch size from around 2k to 4k parallel sentences. We validated every 3k steps calculating chrF on all languages in the training simultaneously, we finished training after 20 validations without improvements. For baselines, it was around 350$\pm$100k steps increasing with corpus size, 4 language models and pivots took around 450$\pm$100k steps and our biggest training run $MultiSlav _{+ENG}$ finished after 880k steps.

\begin{table}[h]
\centering
\begin{tabular}{l| c}
\hline
\textbf{Hyper-parameter} & \textbf{value}\\
\hline
\verb|N encoder layers|& $6$ \\\hline
\verb|N decoder layers|& $6$ \\\hline
$\mathtt{d_{model}}$ & $1024$ \\\hline
$\mathtt{d_{ff}}$ & $4096$ \\\hline
\verb|h|& $16$ \\\hline
\verb|Dropout|& $0.1$ \\\hline
\end{tabular}
\caption{MarianNMT Hyper-parameters used for training models.}
\label{tab:hyper-params-used}
\end{table}

\section{Language Tokens Ablation}
\label{app:lang-tok}
We did not observe any significant difference between the $Many2Many$ models trained with only the target language indicating token or with both the target and the source language indicating tokens. 
$MultiSlav$ model variant with source and target tokens provided final (averaged over all directions) in-training validation\footnote{This score was calculated in-training including added tokens and calculated by MarianNMT chrF implementation, results may differ to unified results reported elsewhere in this study.} \verb|chrF| score of  $52.03$, while single-token counterpart scored \verb|chrF| of $52.14$. 
Therefore, we chose to use the simpler solution, using only the target language tokens.
Special tokens used for indicating each of the target languages are in \autoref{tab:special-lang-toks}.

\begin{table}[h]
\centering
\begin{tabular}{l| c}
\hline
\textbf{Language} & \textbf{Special language token}\\
\hline
\emph{Czech} & \verb|>>ces<<| \\
\emph{English} & \verb|>>eng<<| \\
\emph{Polish} & \verb|>>pol<<| \\
\emph{Slovak} & \verb|>>slk<<| \\
\emph{Slovene} & \verb|>>slv<<| \\\hline
\end{tabular}
\caption{Hyper-parameters used for training models.}
\label{tab:special-lang-toks}
\end{table}

\section{Language specific white-list}
\label{app:white-list}

All of the 5 languages we consider use the Latin script. Our white-list of characters is composed of a `Basic Latin` Unicode block, extended by special characters for each Slavic language supported by a given model. Specific characters added are in \autoref{tab:white-list}.

\begin{table}[h]
\centering
\begin{adjustbox}{max width=0.48\textwidth}
\begin{tabular}{l| l}
\hline
\textbf{Language} & \textbf{White-list characters}=\verb|Basic Latin| + \\
\hline
\emph{Czech} &  áčďéěíňóřšťúůýžÁČĎÉĚÍŇÓŘŠŤÚŮÝŽ \\
% \emph{English} &  \\
\emph{Polish} & ąćęłńóśźżĄĆĘŁŃÓŚŹŻ \\
\emph{Slovak} &  áäčďžéíĺľňóôŕšťúýžÁÄČĎÉÍĹĽŇÓÔŔŠŤÚÝŽ \\
\emph{Slovene} & čćđšžČĆĐŠŽ \\
\hline
\end{tabular}
\end{adjustbox}
\caption{White-list characters}
\label{tab:white-list}
\end{table}

\section{Text Features and Language Identification}
\label{app:text-features-details}

For efficient counting of words in source and target sentences, we used simple tokenization based on splitting on white-spaces. Each split segment of a sentence is considered a word. As a digit, we consider a single string character between 0-9. As a number, we consider a tokenized word containing at least one digit. As a mismatched number, we understand any number in the source sentence that is not present in the target sentence and vice versa. The white-list of characters for each language can be found in \autoref{app:white-list}. For language identification, we use the FastText LangId tool (lid.176 model).
A sentence passes the language detection only if the expected language has the highest LangId probability score. Levenshtein distance was calculated, treating each diacritic as a separate character. The main goal of using Levenshtein distance is to reduce the number of miss-aligned examples in the dataset.
Each filter is applied to source and target sentences separately. See table~\ref{tab:filtering-used} for used values for filtering thresholds.

\begin{table*}
\centering
\begin{tabular}{lll}
\hline
\textbf{Filter} & \textbf{Min Value} & \textbf{Max Value}\\
\hline
\verb|Sent char length| & $5$ & $500$ \\
\verb|Word count| & $1$ & $100$ \\
\verb|Avg word length| & $-$ & $12$ \\
\verb|Max word length| & $-$ & $28$ \\
\verb|Digit ratio| & $-$ & $0.15$ \\
\verb|Non-whitelist ratio| & $-$ & $\leq 0$ \\
\verb|Lang detect| & $0$ & $-$ \\
\verb|Levenshtein Distance| & $2$ & $-$ \\
\verb|Poisson ratio| & $-15.0$ & $-$ \\\hline
\end{tabular}
\caption{Acceptable ranges of filters used for data pre-processing; only examples pair which meet all criteria are chosen; value for each feature must be strictly greater than \emph{Min Value} and strictly lesser than \emph{Max Value}; \emph{Non-whitelist ratio} must be lesser or equal to $0$.}
\label{tab:filtering-used}
\end{table*}

\section{LLM translation refusal}
\label{app:llm-nontranslate}

Flores-101 contains a few sentences that allude to problematic content. Both LLMs that we tested have safety mechanisms preventing the generations of potentially problematic responses. For GPT-3.5 there were 109 empty responses (0.55\%) and for PaLM-2 there were 23 (0.12\%). We manually checked all of them and confirmed that they are associated with issues like: racial stereotyping, sexual content, manslaughter, drugs, or explosives. 

An interesting aspect of our investigation is the fact that the triggering of safety mechanisms was dependent not only on the source sentence content but also on the translation direction. For example, the English sentence 'When the official arrived, the apartment exploded` was correctly translated into Polish but triggered the content filter when translating into Czech. Those aspects might be important when considering a model for a faithful translation of controversial topics, for example, while translating news articles.

\section{Building the tokenizers}
\label{app:tokenizer-sampling}

Before the training, the duplicate sentences were removed from the training set. After deduplication and sampling, each training dataset for tokenizers contained around $40M$ sentences.

\subsection{Tokenizer vocabulary sizes}
\label{app:tokenzier-sizes}

Using the heuristic, that every language should have a comparable amount of tokens within the vocabulary we set their sizes as multiplies of $8k$, $16k$, and $32k$. 
Therefore bi-directional model vocabulary sizes variants were respectively $16k$, $32k$, $64k$. Four language model vocabularies: $32k$, $64k$, $128k$. Five language model vocabularies: $40k$, $80k$, $160k$. The performance results of the models with different tokenizer sizes were similar on automated metrics. Therefore $16k$ per language was chosen as the base size.

\subsection{Data sampling for tokenizer}
Due to the dataset's language imbalance, two different sampling strategies were tested. The first approach used proportional sampling. It preserved the language distribution of the original dataset.
In the second approach, we sampled an equal number of sentences for each language. Regardless of the sampling strategy, the token overlap between their vocabularies was high. It had around 80\% (77.17\%-88.60\%) as long as their vocabulary sizes were proportional to the numbers of supported languages, i.e. 64k four languages tokenizer corresponded to 80k five languages tokenizer. In case the vocabulary sizes differed, the common vocabulary percentage was computed in relation to the smaller one.

The FLoRes-101 dataset was used to evaluate the tokenization effectiveness.
The subsets corresponding to the languages of our interest were tokenized. Their total lengths were calculated with the averages and standard deviations across the languages. The equal data sampling promoted a lower standard deviation and an average length of tokenized sentences.
Regardless of the tokenizer size and used dataset sampling strategy the automated metrics varied only $\pm 0.5 chrF$.

\section{Datasets details}
\label{app:detailed-data-sources}

We trained models on open-source parallel data downloaded via the MTData library. We excluded any datasets which were "for non-commercial use" or "for research-only use". In the \autoref{tab:data-details-app} we named all used corpora.

\begin{table*}
\centering

\begin{tabular}{l r}
\hline
Corpus & Data Size \\\hline
paracrawl & $246407901$ \\\hline
opensubtitles & $167583218$ \\\hline
multiparacrawl & $52388826$ \\\hline
dgt & $36403859$ \\\hline
elrc & $29687222$ \\\hline
xlent & $18375223$ \\\hline
wikititles & $12936394$ \\\hline
wmt & $11074816$ \\\hline
wikimatrix & $10435588$ \\\hline
dcep & $10239150$ \\\hline
ELRC & $7609067$ \\\hline
tildemodel & $6309369$ \\\hline
europarl & $6088362$ \\\hline
eesc & $5604672$ \\\hline
eubookshop & $3732718$ \\\hline
emea & $3482661$ \\\hline
jrc\_acquis & $2920805$ \\\hline
ema & $1881408$ \\\hline
qed & $1835208$ \\\hline
elitr\_eca & $1398536$ \\\hline
EU-dcep & $1132950$ \\\hline
rapid & $1016905$ \\\hline
ecb & $885442$ \\\hline
kde4 & $541944$ \\\hline
news\_commentary & $498432$ \\\hline
kde & $473269$ \\\hline
bible\_uedin & $429692$ \\\hline
europat & $358911$ \\\hline
elra & $357696$ \\\hline
wikipedia & $352118$ \\\hline
wikimedia & $201088$ \\\hline
tatoeba & $91251$ \\\hline
globalvoices & $69736$ \\\hline
euconst & $65507$ \\\hline
ubuntu & $47301$ \\\hline
php & $44031$ \\\hline
ecdc & $21154$ \\\hline
eac & $20224$ \\\hline
eac\_reference & $10099$ \\\hline
gnome & $4466$ \\\hline
EU-eac & $2925$ \\\hline
books & $2816$ \\\hline
EU-ecdc & $2210$ \\\hline
newsdev & $1953$ \\\hline
khresmoi\_summary & $889$ \\\hline
czechtourism & $832$ \\\hline
khresmoi\_summary\_dev & $455$ \\\hline
worldbank & $189$ \\\hline

\end{tabular}
\caption{Corpora used for training, and a respective number of examples, before filtering, deduplication, or any preprocessing.}
\label{tab:data-details-app}
\end{table*}

\section{Detailed results}
\label{app:detailed-results}

In this section we additionally provide BLEU\footnote{Exact version and configuration of BLEU: "nrefs:1 case:mixed eff:no tok:13a smooth:exp version:2.3.1" from SacreBLEU library} metric to chrF and COMET.
Results in the tables below are not averaged and provide results for all models in all supported directions.

\newpage

\begin{table*}
    \begin{adjustbox}{max width=0.65\textheight}
\begin{tabular}{l| c c c c | c c c c | c c c c | c c c c | c c c c}
source language & \multicolumn{4}{c|}{$CES \rightarrow$} & \multicolumn{4}{c|}{$ENG \rightarrow$} & \multicolumn{4}{c|}{$POL \rightarrow$} & \multicolumn{4}{c|}{$SLK \rightarrow$} & \multicolumn{4}{c}{$SLV \rightarrow$} \\
target language & $ENG$ & $POL$ & $SLK$ & $SLV$ & $CES$ & $POL$ & $SLK$ & $SLV$ & $CES$ & $ENG$ & $SLK$ & $SLV$ & $CES$ & $ENG$ & $POL$ & $SLV$ & $CES$ & $ENG$ & $POL$ & $SLK$ \\
\hline
\hline
$GoogleTranslate$ & \underline{88.9} & \underline{91.0} & 92.2 & 90.5 & 91.6 & \underline{90.6} & 91.8 & 90.7 & 91.0 & 86.7 & 90.6 & 89.8 & 92.8 & \underline{89.0} & \underline{91.0} & \underline{90.5} & 91.5 & 88.4 & 90.5 & 91.1 \\
$PaLM-2$ & \underline{88.9} & \underline{91.0} & \underline{92.8} & \underline{90.6} & \underline{92.5} & 90.4 & \underline{92.1} & \underline{91.2} & \underline{91.6} & \underline{86.8} & \underline{91.0} & \underline{90.4} & \underline{93.3} & \underline{89.0} & \underline{91.0} & 89.3 & \underline{92.1} & \underline{88.6} & \underline{90.8} & \underline{91.6} \\
$ChatGPT-3.5$ & 88.3 & 89.8 & 91.9 & 90.2 & 91.0 & 89.5 & 90.1 & 90.1 & 90.6 & 86.2 & 89.6 & 89.3 & 92.9 & 88.2 & 90.0 & 90.3 & 90.4 & 87.9 & 89.8 & 89.6 \\
\hline
$M2M-100$ & 87.0 & 89.0 & 92.1 & 89.7 & 88.6 & 86.4 & 88.4 & 87.3 & 89.6 & 84.6 & 89.4 & 88.4 & 92.7 & 86.8 & 89.1 & 89.6 & 90.3 & 86.4 & 88.7 & 90.1 \\
$NLLB-200$ & 88.1 & 88.9 & 91.2 & 88.6 & 90.4 & \textbf{88.5} & 90.1 & 88.8 & 89.4 & \textbf{85.8} & 88.9 & 87.7 & 91.8 & 88.2 & 88.9 & 88.8 & 90.0 & \textbf{87.5} & 88.6 & 89.4 \\
$Seamless-M4T$ & 87.5 & 80.9 & 90.8 & 82.0 & \textbf{90.7} & \textbf{88.5} & \textbf{90.6} & \textbf{89.6} & 79.6 & 85.4 & 80.0 & 76.4 & 91.5 & 87.2 & 81.2 & 82.9 & 80.9 & 87.3 & 76.7 & 81.0 \\
$OPUS-MT Sla-Sla$ & \textbf{88.2} & 82.8 & - & 83.4 & 89.1 & 85.6 & - & 84.5 & 82.9 & 82.2 & - & 81.2 & - & - & - & - & 83.5 & 84.1 & 80.8 & - \\
$OPUS-MT SK-EN$ & - & - & - & - & - & - & 89.5 & - & - & - & - & - & - & \textbf{88.4} & - & - & - & - & - & - \\
\hline
\hline
Our contribution: &  &  &  &  &  &  &  &  &  &  &  &  &  &  &  &  &  &  &  &  \\
\hline
$baseline$ & 87.5 & 89.4 & 92.4 & 89.8 & 87.8 & 86.2 & 87.2 & 86.6 & 90.0 & 85.0 & 89.1 & 88.4 & 92.9 & 87.3 & 88.8 & 89.4 & 90.0 & 86.9 & 88.1 & 89.1 \\
\hline
$P4_{POL}$ & - & 89.6 & 90.8 & 88.7 & - & - & - & - & 90.2 & - & 89.8 & 88.7 & 91.0 & - & 89.3 & 88.4 & 89.3 & - & 88.7 & 88.5 \\
$P5_{ENG}$ & 88.0 & 89.0 & 90.7 & 89.0 & 88.8 & 87.3 & 88.4 & 87.5 & 89.0 & 85.7 & 88.5 & 87.8 & 91.0 & 88.2 & 88.6 & 88.5 & 89.6 & 87.2 & 88.4 & 88.9 \\
$P5_{CES}$ & 87.9 & 89.6 & \textbf{92.5} & 89.9 & 88.4 & 85.0 & 87.9 & 85.9 & 90.3 & 84.5 & 89.5 & 88.0 & \textbf{93.0} & 87.8 & 89.4 & 89.8 & 90.3 & 85.7 & 87.9 & 89.8 \\
\hline
$MultiSlav$ & - & 89.7 & \textbf{92.5} & 90.0 & - & - & - & - & 90.2 & - & 89.6 & 88.7 & 92.9 & - & 89.4 & 90.1 & 90.6 & - & 88.9 & \textbf{90.2} \\
$MultiSlav_{+ENG}$ & 87.8 & 89.8 & \textbf{92.5} & 90.1 & 88.9 & 86.9 & 88.0 & 87.3 & \textbf{90.4} & 85.4 & 89.8 & \textbf{88.9} & 92.9 & 87.8 & 89.6 & \textbf{90.2} & 90.6 & 87.0 & \textbf{89.2} & \textbf{90.2} \\
\hline
$MultiSlav_{prop}$ & - & 89.7 & \textbf{92.5} & \textbf{90.2} & - & - & - & - & 90.1 & - & 89.6 & 88.8 & 92.9 & - & 89.5 & 90.1 & 90.6 & - & 89.0 & 90.0 \\
$MultiSlav_{exc:sk2sl}$ & - & \textbf{89.9} & \textbf{92.5} & 90.0 & - & - & - & - & 90.3 & - & \textbf{89.9} & 88.8 & 92.9 & - & 89.6 & 90.1 & \textbf{90.7} & - & 89.0 & \textbf{90.2} \\
$MultiSlav_{exc:sl2sk}$ & - & 89.6 & \textbf{92.5} & \textbf{90.2} & - & - & - & - & 90.2 & - & 89.8 & \textbf{88.9} & \textbf{93.0} & - & \textbf{89.7} & \textbf{90.2} & 90.5 & - & 89.0 & 89.9 \\
$MultiSlav_{exc:both}$ & - & 89.6 & 92.4 & 90.1 & - & - & - & - & 90.3 & - & 89.8 & 88.7 & 92.9 & - & 89.6 & 90.0 & 90.6 & - & 88.8 & 89.9 \\
\end{tabular}
\end{adjustbox}
\caption{Detailed \textbf{COMET} results; Higher is better; \underline{Underlined} are the best results, \textbf{bolded} are the best open-source results.}
\label{tab:detailed-results-comet}
\end{table*}

\begin{table*}
\begin{adjustbox}{max width=0.65\textheight}
\begin{tabular}{l| c c c c | c c c c | c c c c | c c c c | c c c c}
source language & \multicolumn{4}{c|}{$CES \rightarrow$} & \multicolumn{4}{c|}{$ENG \rightarrow$} & \multicolumn{4}{c|}{$POL \rightarrow$} & \multicolumn{4}{c|}{$SLK \rightarrow$} & \multicolumn{4}{c}{$SLV \rightarrow$} \\
target language & $ENG$ & $POL$ & $SLK$ & $SLV$ & $CES$ & $POL$ & $SLK$ & $SLV$ & $CES$ & $ENG$ & $SLK$ & $SLV$ & $CES$ & $ENG$ & $POL$ & $SLV$ & $CES$ & $ENG$ & $POL$ & $SLK$ \\
\hline
\hline
$GoogleTranslate$ & \underline{68.2} & \underline{51.6} & 56.1 & \underline{57.0} & \underline{61.8} & \underline{55.1} & \underline{64.7} & 62.0 & 50.1 & \underline{60.8} & \underline{51.6} & 52.3 & 55.9 & \underline{68.6} & 51.2 & \underline{56.9} & 54.2 & \underline{65.6} & \underline{50.7} & \underline{55.5} \\
$PaLM-2$ & 67.2 & 51.5 & \underline{57.9} & 55.3 & 61.8 & 54.5 & 63.5 & \underline{62.2} & \underline{50.2} & \underline{60.8} & 51.2 & \underline{52.8} & \underline{56.8} & 67.6 & \underline{51.4} & 54.2 & \underline{54.7} & 65.3 & \underline{50.7} & 55.1 \\
$ChatGPT-3.5$ & 66.0 & 49.2 & 55.9 & 55.1 & 58.5 & 52.5 & 58.4 & 59.1 & 47.8 & 58.8 & 47.9 & 50.2 & 55.9 & 66.3 & 49.2 & 55.1 & 52.0 & 63.4 & 48.7 & 51.8 \\
\hline
$M2M-100$ & 63.0 & 48.0 & 56.2 & 54.6 & 57.3 & 49.4 & 58.8 & 57.1 & 47.7 & 56.4 & 48.8 & 50.0 & 55.6 & 63.5 & 48.1 & 55.0 & 51.6 & 61.0 & 47.3 & 52.8 \\
$NLLB-200$ & \textbf{65.4} & 47.3 & 54.0 & 51.5 & 57.0 & 50.4 & 59.1 & 56.9 & 46.7 & \textbf{58.6} & 47.5 & 47.8 & 52.9 & 65.9 & 46.8 & 52.0 & 49.9 & \textbf{63.2} & 46.3 & 50.4 \\
$Seamless-M4T$ & 63.4 & 43.5 & 54.2 & 48.3 & 58.5 & 51.3 & 60.1 & \textbf{58.6} & 41.2 & 57.0 & 41.7 & 42.3 & 53.6 & 62.8 & 43.6 & 48.8 & 45.0 & 62.4 & 41.6 & 46.0 \\
$OPUS-MT Sla-Sla$ & 65.6 & 43.5 & - & 48.2 & \textbf{59.5} & 50.0 & - & 54.3 & 42.6 & 53.2 & - & 44.0 & - & - & - & - & 46.0 & 57.8 & 41.8 & - \\
$OPUS-MT SK-EN$ & - & - & - & - & - & - & \textbf{62.1} & - & - & - & - & - & - & \textbf{66.6} & - & - & - & - & - & - \\
\hline
\hline
Our contribution: &  &  &  &  &  &  &  &  &  &  &  &  &  &  &  &  &  &  &  &  \\
\hline
$baseline$ & 64.2 & 49.2 & 56.6 & 55.5 & 58.7 & 50.8 & 60.7 & 58.2 & 48.5 & 56.9 & 49.3 & 50.3 & 56.1 & 64.2 & 48.7 & 55.4 & 52.0 & 61.5 & 47.3 & 52.5 \\
\hline
$P4_{POL}$ & - & \textbf{49.5} & 54.4 & 53.6 & - & - & - & - & 48.5 & - & 49.6 & 50.7 & 53.2 & - & 49.4 & 53.2 & 50.4 & - & \textbf{48.0} & 51.1 \\
$P5_{ENG}$ & 64.9 & 48.7 & 56.0 & 54.8 & \textbf{59.5} & \textbf{51.8} & 61.2 & 58.4 & 48.3 & 58.1 & 49.2 & 50.2 & 54.9 & 65.8 & 48.6 & 54.6 & 52.4 & 62.2 & 47.8 & 52.9 \\
$P5_{CES}$ & 64.7 & 49.0 & 56.7 & 55.3 & 58.9 & 49.2 & 60.6 & 55.9 & 48.6 & 55.5 & 49.0 & 49.5 & 56.2 & 65.0 & 48.9 & 55.3 & 52.4 & 59.2 & 46.7 & 53.0 \\
\hline
$MultiSlav$ & - & 49.2 & 56.7 & \textbf{55.9} & - & - & - & - & 48.7 & - & 49.5 & 50.7 & 56.3 & - & 49.1 & 55.7 & \textbf{52.7} & - & \textbf{48.0} & \textbf{53.6} \\
$MultiSlav_{+ENG}$ & 64.4 & 49.3 & \textbf{56.9} & 55.8 & 59.2 & 51.1 & 60.3 & 58.2 & \textbf{48.9} & 57.4 & \textbf{49.8} & 50.4 & \textbf{56.4} & 64.9 & \textbf{49.5} & 55.7 & 52.5 & 61.7 & 47.9 & 53.3 \\

\hline
$MultiSlav_{prop}$ & - & 49.0 & 56.7 & 55.7 & - & - & - & - & 48.7 & - & 49.4 & 50.7 & 56.2 & - & 49.1 & 55.7 & 52.3 & - & 47.9 & 53.4 \\
$MultiSlav_{exc:sk2sl}$ & - & 49.3 & 56.7 & 55.7 & - & - & - & - & 48.6 & - & 49.6 & 50.7 & 56.2 & - & 49.3 & 55.4 & 52.6 & - & \textbf{48.0} & \textbf{53.6} \\
$MultiSlav_{exc:sl2sk}$ & - & 49.1 & 56.7 & \textbf{55.9} & - & - & - & - & 48.8 & - & 49.6 & \textbf{50.9} & 56.2 & - & 49.3 & \textbf{55.9} & 52.4 & - & 47.8 & 52.9 \\
$MultiSlav_{exc:both}$ & - & 49.2 & 56.7 & 55.6 & - & - & - & - & 48.6 & - & 49.5 & 50.6 & 56.2 & - & 49.2 & 55.5 & 52.3 & - & 47.8 & 53.1 \\
\end{tabular}
\end{adjustbox}
\caption{Detailed \textbf{chrF} results; Higher is better; \underline{Underlined} are the best results, \textbf{bolded} are the best open-source results.}
\label{tab:detailed-results-chrf}
\end{table*}

\begin{table*}
\begin{adjustbox}{max width=0.65\textheight}
\begin{tabular}{l| c c c c | c c c c | c c c c | c c c c | c c c c}
source language & \multicolumn{4}{c|}{$CES \rightarrow$} & \multicolumn{4}{c|}{$ENG \rightarrow$} & \multicolumn{4}{c|}{$POL \rightarrow$} & \multicolumn{4}{c|}{$SLK \rightarrow$} & \multicolumn{4}{c}{$SLV \rightarrow$} \\
target language & $ENG$ & $POL$ & $SLK$ & $SLV$ & $CES$ & $POL$ & $SLK$ & $SLV$ & $CES$ & $ENG$ & $SLK$ & $SLV$ & $CES$ & $ENG$ & $POL$ & $SLV$ & $CES$ & $ENG$ & $POL$ & $SLK$ \\
\hline
\hline
$GoogleTranslate$ & \underline{43.2} & 21.5 & 27.6 & \underline{29.4} & \underline{36.5} & \underline{25.0} & \underline{39.8} & 35.5 & 22.1 & 32.6 & \underline{23.3} & 23.6 & 27.3 & \underline{43.0} & 20.6 & \underline{29.1} & 27.0 & 39.4 & 20.6 & \underline{28.2} \\
$PaLM-2$ & 42.5 & \underline{21.6} & \underline{30.4} & 27.9 & 36.4 & 24.9 & 37.8 & \underline{36.0} & \underline{22.6} & \underline{33.6} & 23.2 & \underline{23.9} & \underline{29.4} & 42.8 &\underline{ 21.2} & 26.7 & \underline{28.3} & \underline{40.3} & \underline{21.0} & 28.0 \\
$ChatGPT-3.5$ & 38.8 & 18.4 & 27.5 & 26.1 & 30.8 & 21.5 & 29.9 & 30.3 & 18.6 & 28.5 & 18.6 & 20.3 & 26.9 & 38.1 & 17.5 & 26.0 & 23.9 & 35.1 & 18.0 & 23.0 \\
\hline
$M2M-100$ & 36.7 & 18.3 & 28.0 & 26.0 & 30.4 & 19.0 & 32.0 & 29.3 & 19.8 & 27.5 & 20.8 & 20.9 & 27.6 & 36.7 & 17.6 & 26.6 & 24.3 & 33.9 & 17.8 & 24.9 \\
$NLLB-200$ & \textbf{40.1} & 17.5 & 26.3 & 23.0 & 30.0 & 20.1 & 32.2 & 29.2 & 19.2 & \textbf{30.9} & 19.2 & 18.6 & 24.9 & 40.3 & 17.0 & 23.3 & 22.5 & \textbf{37.6} & 17.0 & 22.1 \\
$Seamless-M4T$ & 37.7 & 11.6 & 24.5 & 16.9 & 30.8 & 20.6 & 32.8 & \textbf{31.0 }& 11.6 & 29.3 & 11.8 & 10.9 & 23.3 & 37.0 & 11.8 & 17.5 & 14.9 & 36.4 & 10.0 & 15.9 \\
$OPUS-MT Sla-Sla$ & 39.8 & 13.6 & - & 18.6 & \textbf{32.8} & 19.5 & - & 24.8 & 14.6 & 22.9 & - & 14.8 & - & - & - & - & 17.4 & 28.7 & 12.3 & - \\
$OPUS-MT SK-EN$ & - & - & - & - & - & - & \textbf{36.0} & - & - & - & - & - & - & \textbf{40.4} & - & - & - & - & - & - \\
\hline
\hline
Our contribution: &  &  &  &  &  &  &  &  &  &  &  &  &  &  &  &  &  &  &  &  \\
\hline
$baseline$ & 36.9 & 19.2 & 28.2 & 27.0 & 31.4 & 20.1 & 34.1 & 29.8 & 20.3 & 26.7 & 20.6 & 20.9 & 27.7 & 36.2 & 18.0 & 26.6 & 24.4 & 33.3 & 17.0 & 24.3 \\
\hline
$P4_{POL}$ & - & \textbf{19.3} & 27.1 & 24.9 & - & - & - & - & 20.2 & - & 20.9 & 21.5 & 25.5 & - & \textbf{19.1} & 24.8 & 22.7 & - & 17.7 & 23.3 \\
$P5_{ENG}$ & 37.9 & 18.5 & 28.2 & 25.7 & 32.6 & \textbf{21.0} & 34.8 & 30.0 & 19.9 & 28.8 & 20.5 & 20.5 & 26.9 & 38.6 & 18.2 & 25.6 & 24.7 & 34.8 & 17.6 & 24.6 \\
$P5_{CES}$ & 37.9 & 19.0 & 28.1 & 26.7 & 31.9 & 18.8 & 33.7 & 27.4 & 20.2 & 26.2 & 20.4 & 20.0 & 27.8 & 37.9 & 18.3 & 26.6 & 24.9 & 31.4 & 16.7 & 24.9 \\
\hline
$MultiSlav$ & - & 18.9 & 28.3 & \textbf{27.5} & - & - & - & - & 20.6 & - & 21.2 & 21.6 & 27.9 & - & 18.7 & 27.1 & \textbf{25.3} & - & 17.8 & \textbf{25.8} \\
$MultiSlav_{+ENG}$ & 37.2 & 19.0 & \textbf{28.6} & 27.4 & 32.0 & 20.5 & 33.4 & 30.2 & \textbf{20.7} & 27.7 & \textbf{21.3} & 21.2 & \textbf{28.0} & 37.2 & 18.7 & 27.1 & 25.2 & 33.7 & \textbf{18.0} & 25.3 \\
\hline
$MultiSlav_{prop}$ & - & \textbf{19.3} & 28.3 & 27.3 & - & - & - & - & 20.3 & - & 20.7 & 21.4 & 27.6 & - & 18.8 & 27.2 & 24.7 & - & 17.6 & 25.4 \\
$MultiSlav_{exc:sk2sl}$ & - & \textbf{19.3} & 28.5 & 27.1 & - & - & - & - & 20.4 & - & 20.9 & 21.5 & 27.6 & - & 18.9 & 26.6 & 24.9 & - & \textbf{18.0} & \textbf{25.8} \\
$MultiSlav_{exc:sl2sk}$ & - & 18.9 & 28.4 & 27.4 & - & - & - & - & 20.6 & - & 21.2 & \textbf{21.9} & 27.7 & - & 18.8 & \textbf{27.4} & 24.6 & - & 17.8 & 25.0 \\
$MultiSlav_{exc:both}$ & - & 19.1 & 28.3 & 27.0 & - & - & - & - & 20.4 & - & 21.0 & 21.2 & 27.7 & - & 18.7 & 26.8 & 24.8 & - & 17.8 & 25.2 \\
\end{tabular}
\end{adjustbox}
\caption{Detailed \textbf{BLEU} results; Higher is better; \underline{Underlined} are the best results, \textbf{bolded} are the best open-source results.}
\label{tab:detailed-results-bleu}
\end{table*}

\end{document}